\pdfoutput=1
\documentclass[10pt,twocolumn,letterpaper]{article}

\usepackage[pagenumbers]{cvpr} 

\usepackage{graphicx}
\usepackage{amsmath}
\usepackage{amssymb}
\usepackage{booktabs}

\usepackage{multirow}
\usepackage{makecell}
\usepackage{color}
\newcommand{\Reffig}[1]{Figure~\ref{#1}}

\newcommand{\Refeq}[1]{Equation~\ref{#1}}
\newcommand{\Reftab}[1]{Table~\ref{#1}}

\usepackage[pagebackref,breaklinks,colorlinks]{hyperref}

\usepackage[capitalize]{cleveref}
\crefname{section}{Sec.}{Secs.}
\Crefname{section}{Section}{Sections}
\Crefname{table}{Table}{Tables}
\crefname{table}{Tab.}{Tabs.}

\def\cvprPaperID{8128} 
\def\confName{CVPR}
\def\confYear{2022}

\begin{document}

\title{
Patch-NetVLAD+: Learned patch descriptor and weighted matching strategy \\ for place recognition
}

\author{{Yingfeng Cai}\\
{\tt\small caiyingfeng@tongji.edu.cn}
\and
Junqiao Zhao\\
{\tt\small zhaojunqiao@tongji.edu.cn}
\and
Jiafeng Cui\\
{\tt\small Jiafengcui@tongji.edu.cn}
}
\maketitle

\begin{abstract}
Visual Place Recognition (VPR) in areas with similar scenes such as urban or indoor scenarios is a major challenge. 
Existing VPR methods using global descriptors have difficulty capturing local specific regions (LSR) in the scene and are therefore prone to localization confusion in such scenarios.
As a result, finding the LSR that are critical for location recognition becomes key.
To address this challenge, we introduced Patch-NetVLAD+, which was inspired by patch-based VPR researches.
Our method proposed a fine-tuning strategy with triplet loss to make NetVLAD suitable for extracting patch-level descriptors.
Moreover, unlike existing methods that treat all patches in an image equally, our method extracts patches of LSR, which present less frequently throughout the dataset, and makes them play an important role in VPR by assigning proper weights to them. 
Experiments on Pittsburgh30k and Tokyo247 datasets show that our approach achieved up to 6.35\% performance improvement than existing patch-based methods. 
\end{abstract}

\section{Introduction}
\label{sec:intro}
Visual Place Recognition (VPR) is the task to estimate the location of a query image by recognizing the same place in a set of database images. 
VPR has achieved significant progress with the advances of convolutional neural networks (CNN) 
\cite{arandjelovic2016NetVLAD,garg2018lost,ge2020self,sarlin2019coarse}.
However, VPR is still a challenging task in areas with similar scenes such as urban or indoor scenes.
There are two common ways to address this challenge: improving the performance of descriptor or improving the match strategy.
\begin{figure}
  \centering
  \includegraphics[width=\linewidth]{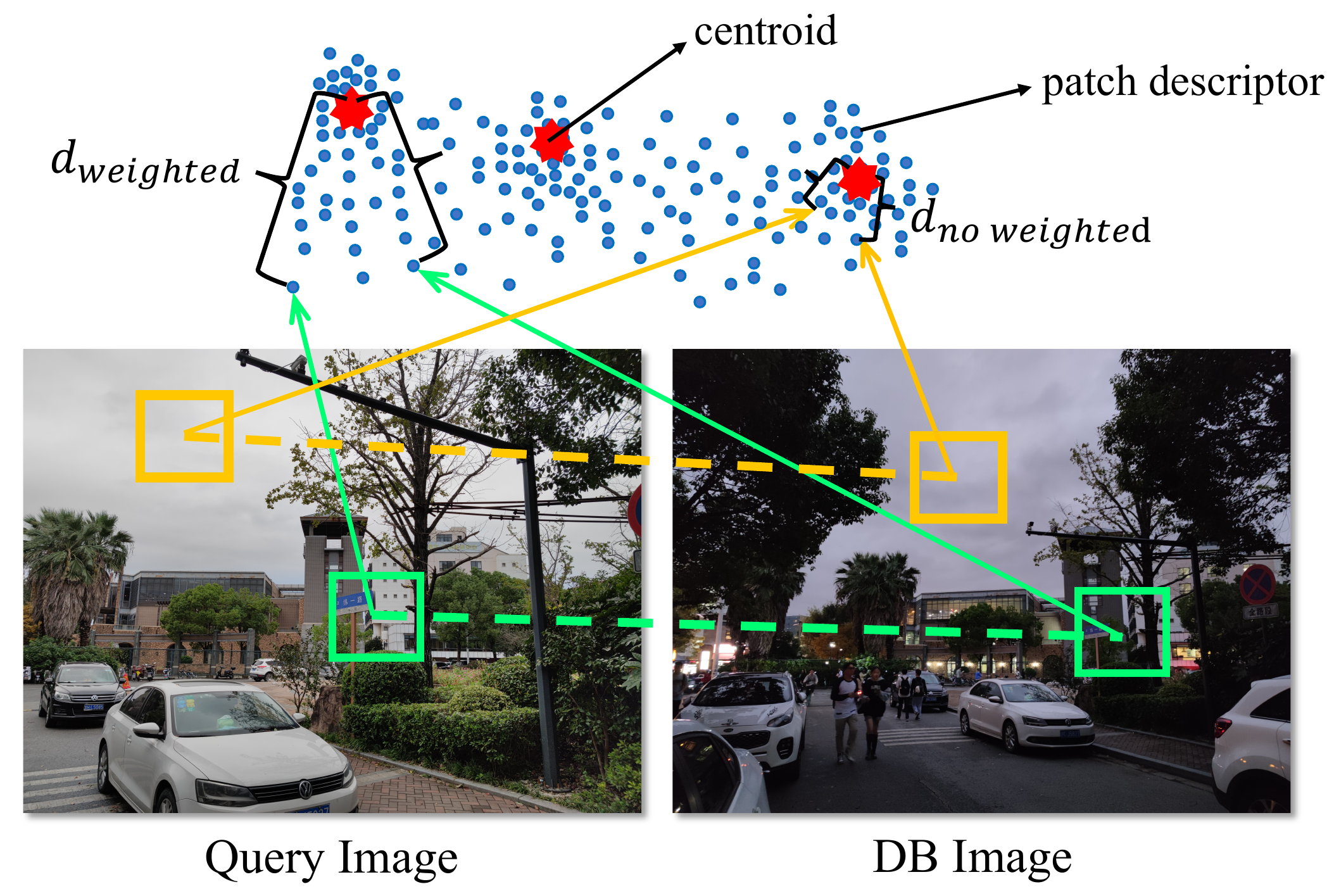}
  \caption{
  \textbf{Comparing weighted and unweighted matching.}
  The yellow patches located in the sky are frequently present in scenes. 
  Therefore, they are depicted close to a centroid (red) of the description space. 
  Therefore they should play an insignificant role in the VPR. 
  On the contrary, patches in LSR (\eg green patches covering the signage) appear only in a few scenes.
  Their descriptors are far from the centroids. 
  We propose to assign weights to patches based on the distance between patch descriptors and centroids in the description space so that patches in LSR play an important role in image matching.
  }
  \label{fig_intro}
\end{figure}

Most of the methods to improve descriptor performance design a novel CNN for extracting global descriptors.
These global descriptors\cite{arandjelovic2016NetVLAD, garg2018lost, garg2019semantic} which describe the whole image typically excel in terms of their robustness to appearance and illumination changes, as they are directly optimized for place recognition. 
However, as shown in \cite{quteprints84931,hausler2021patch, zhu2020regional}, in areas with similar scenes, the global descriptor has difficulty distinguishing differences between local regions.

Our method takes the same strategy as Patch-NetVLAD\cite{hausler2021patch}, which uses a sliding window to generate patches and then derives patch descriptors from NetVLAD to achieve place recognition. 
However, Patch-NetVLAD extracted patch descriptors using NetVLAD trained on the whole image, which is not accurate.
Inspired by SFRS\cite{ge2020self}, our method uses triplet loss fine-tune NetVLAD to make it more suitable for extracting patch-level descriptors.
However, GNSS-based labels of public datasets only provide image-to-image similarities. 
To tackle this problem, patch-to-patch similarities and matches of local keypoint are used to select the positive and negative patch training set required by the triplet loss function.

The methods to optimize the matching strategy are mainly based on intra-set similarities\cite{neubert2021resolving, Schubert2021beyondann} and sequence searching \cite{garg2021seqnet, milford2012seqslam}.
Unlike patch-based methods, these methods optimized the VPR by matching multiple query images, rather than improving the matching performance of a single image.
  Besides, these methods required multiple query images.
The patch-based VPR method decomposed the image into patches, and therefore, utilized the local regions for matching.
Rather than treating all the patches equally, we found that the importance of different patches in pairwise image matching varies.
Therefore, it is critical to find patches in local specific regions (LSR) that are crucial for place recognition.

We propose to cluster the patch descriptors of the whole dataset in the description space, and patches far away from centroids are considered as patches in LSR as they occur less frequently in the dataset.
As shown in \Reffig{fig_intro}, patches located in the frequently present sky are closer to a centroid while patches covering the signage which appear infrequently are far from the centroids.
Inspired by this, by assigning a greater weight to patches in LSR than other patches, we make them play an important role in image matching.


In summary, our contributions are summarised as follows:

1. We propose a novel method to fine-tune NetVLAD using triplet loss to make it more suitable for extracting patch-level descriptors. 
The positive and negative patches of triplet loss function are selected by patch-to-patch similarities and matches of local keypoint.

2. We propose to assign weights to patches in LSR based on distances of patch descriptors from the centroids in the description space throughout the dataset.
The weighted patches are then used to optimize the pairwise image matching.


\section{Related Work}
Existing works on VPR enhancement can be grouped into descriptor-centered and strategy-centered, where the former improved the accuracy and robustness of VPR by designing a novel descriptor, while the works based on the strategy utilized inter-image relationships to optimize VPR.

\subsection{Descriptor-centered Methods}


\textbf{Global Image Descriptor:}
Early global descriptor approaches focused on statistical information of the image, such as color histograms\cite{ulrich2000appearance} and linear image features\cite{krose2001probabilistic}.
Global image descriptor can be aggregated from the local keypoint descriptor, \eg WI-SURF\cite{badino2012real}, ERIEF-Gist\cite{sunderhauf2011brief}, Bag of Words (BoW)\cite{jegou2011aggregating}, Vector of Locally Aggregated Descriptors (VLAD)\cite{{jegou2010aggregating}} and Fisher Vectors (FV)\cite{sanchez2013image}, etc.
Inspired by VLAD, \cite{arandjelovic2016NetVLAD} designed an end-to-end (CNN) architecture NetVLAD that aggregates local descriptors according to learnable centroids.
Benefiting from the success of NetVLAD, Contextual Reweighting Network (CRN)\cite{kim2017crn}, added a spatial attention mechanism before the VLAD layer to focus on regions that positively contribute to VPR.
For the same purpose as CRN, Attention-based Pyramid Aggregation Network (APANet)\cite{zhu2018attention} encode the multi-size buildings by using spatial pyramid pooling, and suppress the confusing region while highlight the discriminative region by attention block.
However, both CRN and APANet require re-training on the whole dataset. 

In order to deal with the problem caused by the opposite viewpoint, \cite{garg2018lost} proposed a novel image descriptor by combining semantic labels and appearance.
However, \cite{garg2018lost} just encodes semantic regions in images by \cite{lin2017refinenet} and simply concatenates them as global descriptors.
In addition, only three types of semantic regions were considered, therefore, it is not valid for images that do not contain these regions.

\textbf{Patch/Region Descriptor:}
This kind of methods computes image descriptors only using relevant patches of an image rather than the whole image.
These methods focus on the regions of interest (\eg, landmarks) of an image.
Most of these methods adopt existing object proposal techniques to extract regions or patches.
\cite{quteprints84931} proposed a landmark-based VPR method by combining the CNN features of landmarks detected by Edge Box\cite{zitnick2014edge}. 
Unlike \cite{quteprints84931}, \cite{chen2017only} only used CNN once to detect landmarks and extract their features.
\cite{garg2018lost} retrained RefineNet\cite{lin2017refinenet} to get semantic regions (\eg, road, building and vegetation) and their descriptors (LoST).

\cite{yang2019landmark} proposed a new landmark generation method named MSW (multi-scale sliding window). 
MSW performed better than object proposal techniques due to the use of the sliding window, especially when illumination or viewpoint changes.
Similar to MSW, Patch-NetVLAD\cite{hausler2021patch} also used the sliding window to generate multi-scale patches and obtained patch descriptors from NetVLAD residuals.
However, these patches are described using NetVLAD trained on the whole image, which is not accurate.

In a general sense, for similar scenes, though the images are similar as a whole, there may still be some dissimilar regions in the images.
Therefore, we also use a sliding window to acquire patches but fine-tune NetVLAD to make it more suitable for patch descriptor extraction.

%

\subsection{Strategy-centered Methods}
SeqSLAM\cite{milford2012seqslam} changed the VPR problem from calculating the single location globally to finding a local best candidate within each local sequence.
Similar ideas were proposed in\cite{hansen2014visual, neubert2019neurologically}.
However, most of these sequence-based methods are designed to improve the performance of VPR rather than to guide the single image to match correctly.
\cite{neubert2021resolving} optimized image matching by considering the relationships within the query dataset; in detail, it inhibits multiple query descriptors that are mutually different from matching to the same database descriptor, but the method is no longer valid when there is only one query image.
\cite{zhu2020regional} proposed a regional relation module to model the relation information between regions of an image, although this method considers the importance of different regions in the image, the extraction and importance of regions are included in the network training, which is closely related to the training set with poor generalization performance, while our method is explicit and has little related to the training set.

Patch-NetVLAD\cite{hausler2021patch} first found the candidate images using NetVLAD, and then calculated the patch match score for each pair of images to rank the candidate images and determine the best matching image.
However, Patch-NetVLAD treated all patches equally when matching patches. 

Different from Patch-NetVLAD, our method tried to find patches in LSR and differentiated patches according to their contributions to VPR. 

\begin{figure*}
  \centering
  \includegraphics[width=0.9\textwidth]{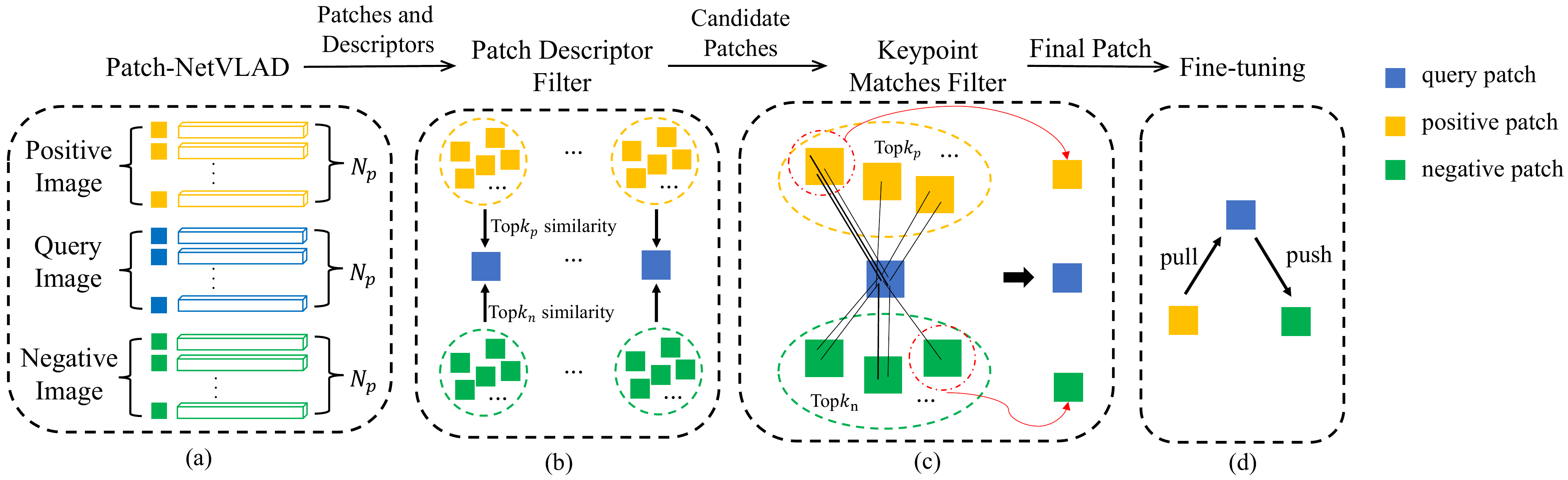}
  \caption{\textbf{The pipeline of fine-tuning the NetVLAD for patches.} 
  (a) The patches are extracted in the images using a sliding window; (b) For each query patch we find $Top\ k_p$ similar patches in the positive image using the patch descriptors extracted from the original NetVLAD, and use these patches as the candidate positive patches; the candidate negative patches are the $Top\ k_n$ similar patches found from the negative image.
  (c) After that, the final positive patch and negative patches are chosen from candidate patches based on the result of keypoint matching.
  For clarity, a query image only selects one positive image and one negative image in the figure.
  (d) The triplet loss-based fine-tuning results in pulling the descriptors of positive patches and query patch together while pushing the descriptors of negative patches and query patch apart.
  }
  \label{fig_finetuning}
\end{figure*}

\section{Methodology}
In this section, we will introduce our method in detail.
Our approach consists of two parts, the first part describes the extraction of patch-level descriptors; the second part describes how to find and assign weights to patches based on their descriptors, and how to make these weighted patches optimize the matching of image pairs.


The notations we will use throughout the paper are as follows:
we denote $P_{i}^{source}$ as a patch $i$, and its descriptor $f_{i}^{source}$, where $source$ can be $db$ the entire database set, $q$ the query image, $p$ the positive image, $n$ the negative image, and $r$ the reference image.
$N_p^{source}$ represents the number of patches in the $source$.

\subsection{Patch-level Descriptor}
\subsubsection{Patch and patch descriptor}

Similar to \cite{hausler2021patch}, we use a sliding window to extract square patches from the feature map (conv5 layer for VGG16) $F \in\mathbb{R} ^{H\times W\times D} $. 
For each image, we can obtain a total of $N_p$ patches$\{P_i \in \mathbb{R}^{(d_p\times d_p)\times D} \}_{i=1}^{N_p}$, and
\begin{equation}
  N_p = \left\lfloor \frac{H-d_p}{s_p}+1 \right\rfloor \times \left\lfloor \frac{W-d_p}{s_p}+1 \right\rfloor, d_p\leqslant min(H,W)
\end{equation}
where $d_p$ is the side length of the square patch, $s_p$ is the stride of the sliding window.

For each patch, we extract its descriptor $\{f_i \in \mathbb{R}^{1\times D_{pca}} \}_{i=1}^{N_p}$ by using a VLAD aggregation layer followed by a projection layer and principle component analysis (PCA).
Concretely, the VLAD aggregation layer $l_{VLAD}$ aggregates the $(d_p\times d_p)\times D$-dimensional features into a $K\times D$-dimensional matrix by applying a weighted sum soft-assignment to the residuals between each feature and K learned centroids. 
Then, the projection layer $l_{proj}$ reshapes the resultant matrix into a vector by applying intra-normalization in its column followed by L2-normalization in its whole. 
And finally, PCA $l_{pca}$ with whitening is applied to reduce the dimension of the output descriptor.
More details can be found in \cite{arandjelovic2016NetVLAD} \cite{hausler2021patch}.

\subsubsection{The selection of positive and negative patches}

Accurate extraction of positive and negative patches is the key to patch descriptor fine-tuning, and the pipeline is shown in \Reffig{fig_finetuning} (a)-(c).


We first use the weak GPS labels provided by the dataset to find the positive image and the negative image. 
Subsequently, patches $\{P_i^q \}_{i=1}^{N_p}$ ($\{P_i^p \}_{i=1}^{N_p}$,$\{P_i^n \}_{i=1}^{N_p}$) and their descriptors$\{f_i^q\}_{i=1}^{N_p}$ ($\{f_i^p\}_{i=1}^{N_p}$,$\{f_i^n\}_{i=1}^{N_p}$) are extracted from the query image, positive image, negative image respectively as labels for training (\Reffig{fig_finetuning} (a)). 

As shown in \Reffig{fig_finetuning} (b), each patch located in the query image is regarded as a candidate query patch $\{P_i^{q} \}_{i=1}^{N_p}$.
For each candidate query patch, the most $k_{p}$ similar patches from the positive image are selected as the candidate positive patches $\{P_i^{p} \}_{i=1}^{k_{p}}$ using the Cosine distance of their descriptors.
Similarly, the most $k_{n}$ similar patches from the negative image are selected as the candidate negative patches $\{P_i^{n} \}_{i=1}^{k_{n}}$.

However, the selected candidate patches are not accurate yet because the descriptors were derived from the original NetVLAD.
Therefore, we extract dense keypoints \cite{Luo_2020_ASLFeat} in the images and match between the query image and its corresponding positive and negative candidate images, followed by GCRANSAC\cite{GCRansac2018} filtering.
Then, we keep the patches with matching keypoints as the final query patches $\{P_i^{q\_f} \}$.
The one with the most matching keypoints from the candidate positive patches is selected as the final positive patch $\{P_i^{p\_f} \}$, and the one with the least number of matched keypoints from the candidate negative patches is selected as the final negative patches $\{P_i^{n\_f} \}$.
This is demonstrated in \Reffig{fig_finetuning} (c). 

\subsubsection{Fine-tune the NetVLAD}

The query patch $\{P_i^{q\_f} \}$ and the corresponding positive patch $\{P_i^{p\_f} \}$ and negative patches $\{P_i^{n\_f} \}$ are used to fine-tune the original NetVLAD with a triplet loss.
Triplet loss for patches is defined as:
\begin{equation}
\begin{aligned}
  loss=\sum_jl\bigg(\min_i\big(d_{cos}(P^q,P_i^p)\big) +m-d_{cos}(P^q,P_j^n)\bigg)
\end{aligned}
\end{equation}
where $l$ is the hinge loss $l(x)=\max(x,0)$, and $m$ is a constant parameter giving the margin.

The triplet loss pulls the positive patches and query patch together, \ie reducing their Cosine distance, while pushes the negative patches and query patch apart, \ie increase their Cosine distances (\Reffig{fig_finetuning} (d)).

\subsection{Weighted patch-based matching}



\subsubsection{Patch weightings}

We aim to find patches in LSR and assign weights to patches according to their contributions in VPR.
A fact is that patches with few occurrences in the DB can be considered as special patches, and a large weight can be assigned to these patches to make them play a major role in VPR.
However, even with a fine-tuned descriptor, it is not able to assess the occurrence of a patch.
Therefore, our approach measures patch specificity by computing the distance between the patch and the dataset, and more specifically, the distance between the patch and the centroids of the DB dataset.

We first extract a patch descriptors set $\{f_i^{db} \in \mathbb{R}^{1\times D_{pca}}\}_{i=1}^{N_p^{db}}$ from the images of the entire DB dataset.
Then, K-means is employed to cluster these patch descriptors to obtain K centroids $\{f_i^{c} \in \mathbb{R}^{1\times D_{pca}}\} _{i=1}^{k}$, and these centroids represent the distribution of similar patch descriptors in the description space.

The distances between a patch descriptor and centroids can be computed using the Cosine distance.
The patches that far away from most of the centroids are considered as special patch, so we define the weighting as following:
\begin{equation}
  \label{eq_weighting}
  w(f)=\sum_i^{\alpha}\{d_{cos}(f,f_i^{c})\}_{\min_{\alpha}}
\end{equation}
where $f$ is the descriptor of the patch that needs to be weighted, $\{ \}_{\min_{\alpha}}$ represent a subset of $\alpha$ smallest items.

\begin{figure*}
  \centering
  \includegraphics[width=0.9\textwidth]{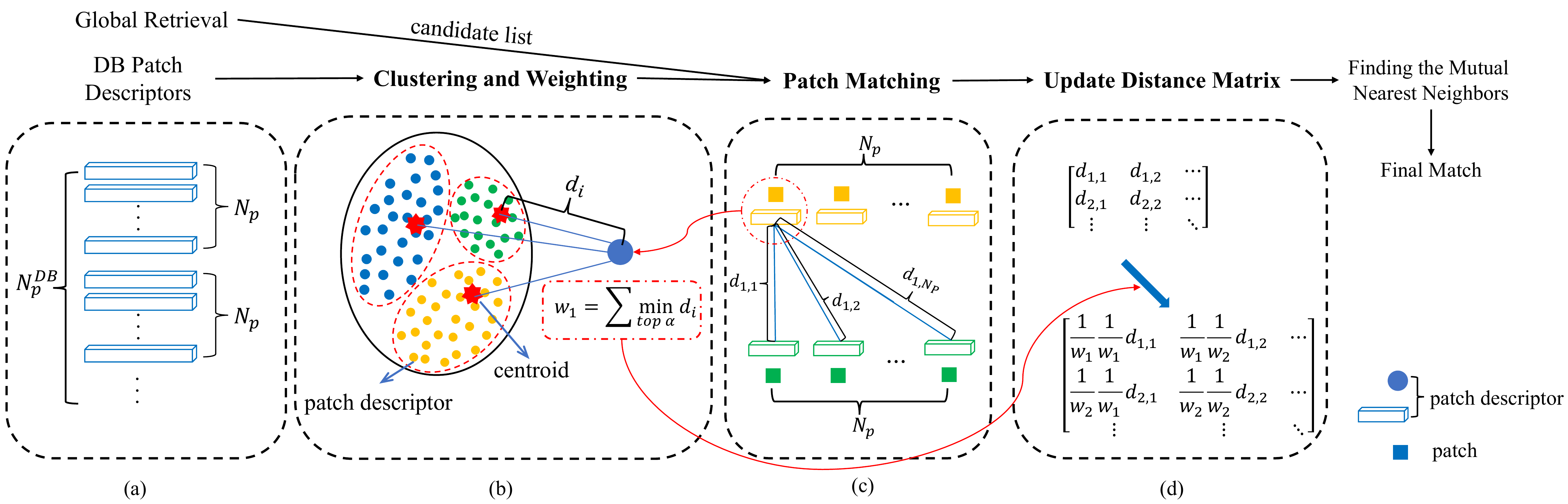}
  \caption{\textbf{The weighted matching strategy.}
  (a)(b) Using K-means to cluster the patch descriptors of the whole DB dataset, patches are weighted according to the distances between the patch descriptors and centroids.
  (c) The distance matrix is derived by exhaustively matching the patches of image pairs.
  (d) The distance matrix is updated by multiplying the weights of the corresponding patches, the mutual nearest neighbors \Refeq{eq_mutual_nn} is optimized based on the updated distance matrix.
  }
  \label{fig_weighting}
\end{figure*}

\subsubsection{Image pair matching }

\cite{hausler2021patch} proposed a hierarchical strategy to find the best matching image of a query image.
First, it used the original NetVLAD description to retrieve the top100 images that are most similar to the query image.
Then, it computed the patch descriptors and perform patch-level matching by finding the mutual nearest neighbors $\mathcal{P}$ by \Refeq{eq_mutual_nn}:
\begin{equation}
  \label{eq_mutual_nn}
  \mathcal{P}=\{(i,j):i=NN_r(f_j^q),j=NN_q(f_i^r)\}
\end{equation}
where $ NN_r(f)=argmin_i\big(d_{cos}(f,f_i^r)\big) $ and $ NN_q(f)=argmin_j\big(d_{cos}(f,f_j^q)\big)$ retrieve the nearest neighbors descriptor match with respect to Cosine distance within the query and reference image.
According to the matched patches, the spatial matching score was computed to rank the top100 images, and the final image retrieval results are obtained.

Similarly, our method also ranks the initial retrieval image set by a similarity score between a pair of images.

Let the image pair list $L$ be:
\begin{equation}
  L={(I_q,I_r)}
\end{equation}
where $I_q$ is the query image, $I_r$ is a candidate image obtained by original NetVLAD.
For each image pair in $L$, the patch descriptors are extracted and their distances are calculated to generate the distance matrix $\mathcal{D}$:
\begin{equation}
  \mathcal{D} =
  \begin{bmatrix}
    d_{cos}(f_1^q,f_1^r)&d_{cos}(f_1^q,f_2^r)&\cdots  \\
    d_{cos}(f_2^q,f_1^r)&d_{cos}(f_2^q,f_2^r)&\cdots  \\
    \vdots &\vdots &\ddots  
  \end{bmatrix}
\end{equation}

In order to make the higher weighted patches located in the LSR play a bigger role in matching, we weighted the distance matrix to update $\mathcal{D}$. 
The weighted distance matrix is given by:
\begin{equation}
  weighted\ \mathcal{D} = \\
  \begin{bmatrix}
    \frac{1}{w(f_1^q)} \frac{1}{w(f_1^r)} &\frac{1}{w(f_1^q)} \frac{1}{w(f_2^r)} &\cdots  \\
    \frac{1}{w(f_2^q)} \frac{1}{w(f_1^r)} &\frac{1}{w(f_1^q)} \frac{1}{w(f_2^r)}&\cdots  \\
    \vdots &\vdots &\ddots 
  \end{bmatrix}
  \circ \mathcal{D} 
\end{equation}
where $\circ $ means Hadamard product.
By this means, patches with large weights can be matched easily, because their values in the distance matrix become smaller, while those with small weights are hardly matched.


\section{Experiments}

\subsection{Experiment setup}

We use two benchmark datasets to evaluate our method: Pittsburgh30k \cite{torii2013pitts} and Tokyo247 \cite{torii201524tokyo247}.
Both datasets are captured at large-scale scenes that contain many similar scenes.
Pittsburgh30k benchmark consists of three parts: train-set, val-set and test-set.
The train-set contains 7416 queries and 10000 reference images; 
the val-set contains 7608 queries and 10000 reference images;
the test-set contains 6818 queries and 10000 reference images.
Tokyo247 consists of 315 queries and  75984 reference images.
For a fair comparison, we follow the experiment settings of \cite{hausler2021patch} and choose same queries as \cite{hausler2021patch} of Pittsburgh30k and Tokyo247.
In our experiment, we adopt the square patch with side length \textbf{$d_p=5$} on feature map (the output of conv5 layer for VGG16) and \emph{Rapid Spatial Scoring} proposed by \cite{hausler2021patch}.

During fine-tuning the NetVLAD, the training parameters are the same as \cite{arandjelovic2016NetVLAD}.
We utilize the train-set of Pittsburgh30k for fine-tuning the original NetVLAD and the best model that achieves optimal performance on val-set is selected.

The original NetVLAD is used to extract the patch descriptors to find the candidate positive patches and negative patches.
ASLFeat\cite{Luo_2020_ASLFeat} is used to detect the keypoints on the image, after detecting keypoints, KNN match ($k=2$) is used to match keypoints followed by GCRANSAC\cite{GCRansac2018} to remove the outlier match.

During weighted matching, we set the stride of the sliding window to $s_p=d_p=5$, \ie there is no overlap between patches, the number of centroids is set to $k=16$ and $\alpha =10$ in \Refeq{eq_weighting}.

All datasets are evaluated using Recall@N metric, \ie a query image is successfully retrieved from $top N$ if at least one of the $top N$ images is within 25 meters.

\subsection{Comparison with the State-of-the-art}
\setlength{\tabcolsep}{14pt} 
\renewcommand\arraystretch{1.1}
\begin{table*}[tbp]
  \small
  \begin{center}
      \caption{Quantitative results} 
      \label{tb_quantitative}
      \begin{tabular}{@{}c|ccc||ccc@{}}
      \toprule
      \multirow{2}{*}{Method}   & \multicolumn{3}{c||}{Pittsburgh30k} & \multicolumn{3}{c}{Tokyo247} \\ \cmidrule(l){2-7} 
                                              & R@1         & R@5        & R@10        & R@1     & R@5    & R@10  \\ \midrule \midrule 
      NetVLAD\cite{arandjelovic2016NetVLAD}   & 83.66       & 91.77      & 94.04       & 66.89   & 78.10  & 80.95 \\
      Patch-NetVLAD\cite{hausler2021patch}    & 87.40       & 93.66      & 95.36       & 77.14   & 84.44  & 86.98 \\ \hline
      Ours                                    & 88.50       & 94.47      & 95.79       & 83.49   & 87.94  & 90.16 \\ \bottomrule
      
      \end{tabular}
  \end{center}
  \vspace{-0.3cm}
\end{table*}


\begin{figure*}[t]
  \centering
      $\begin{array} {cc}
      \includegraphics[width=3in]{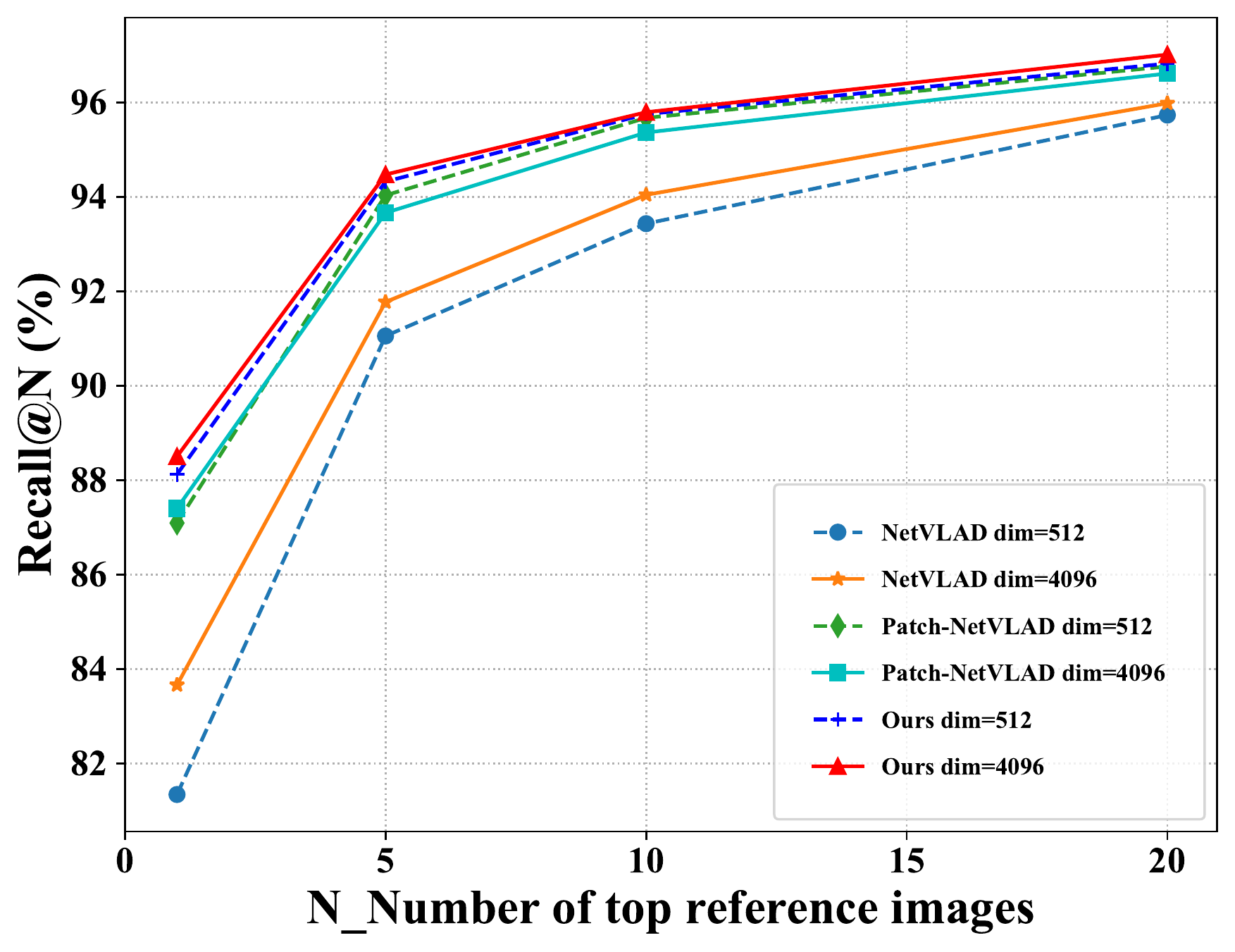} &
      \includegraphics[width=3in]{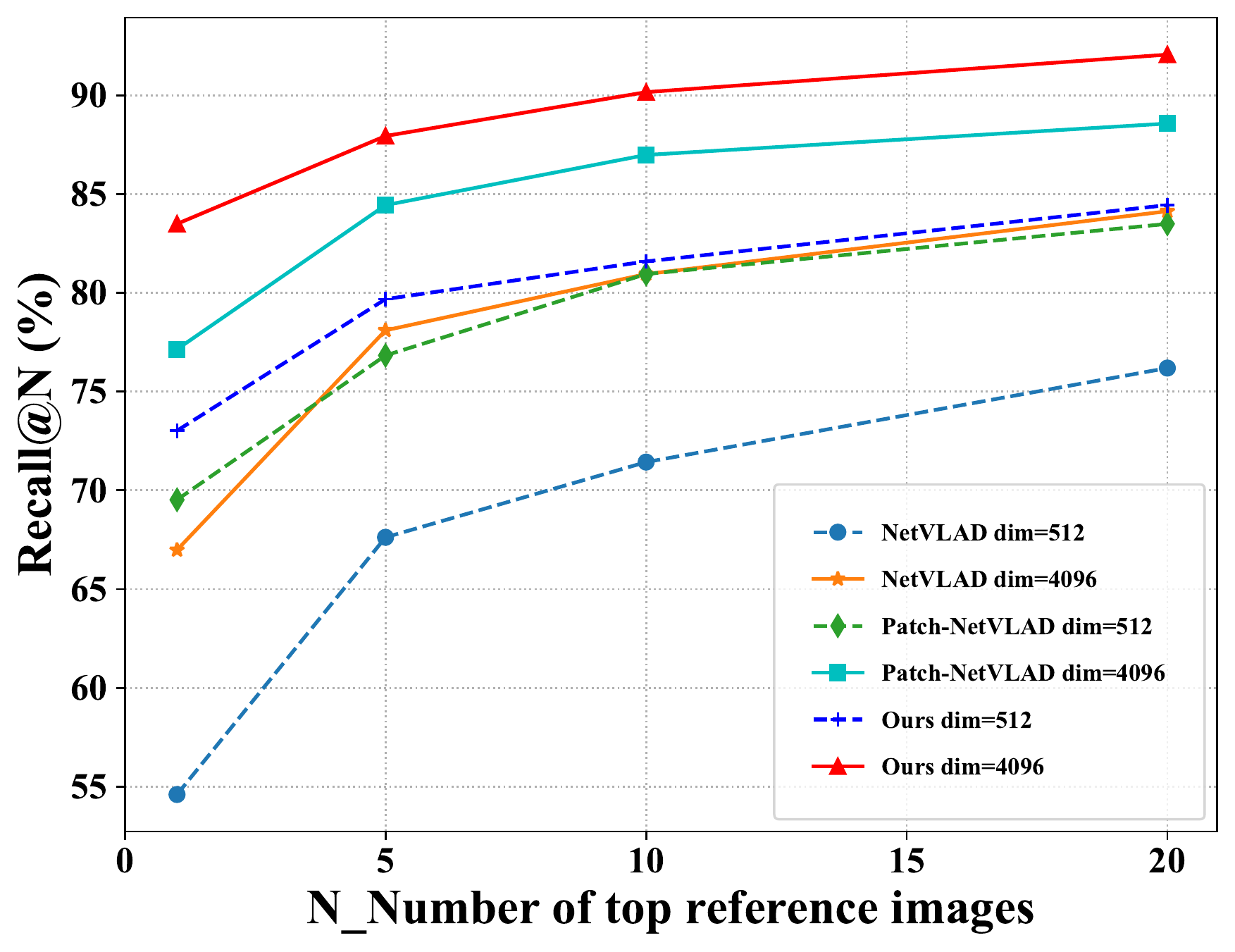} \\
      \mbox{(a) Pittsburgh30k} &\mbox{(b) Tokyo247}
      \end{array}$
  \caption{
  \textbf{Comparison with the State-of-the-art.} We show the comparison of the Recall@N performances with state-of-the-art methods NetVLAD\cite{arandjelovic2016NetVLAD} and Patch-NetVLAD\cite{hausler2021patch} on Pittsburgh30k and Tokyo247. Our method is only fine-tuned on Pittsburgh30k.
  }
  \label{fig_compare}
\end{figure*}

\begin{figure*}
  \centering
  \includegraphics[width=0.9\textwidth]{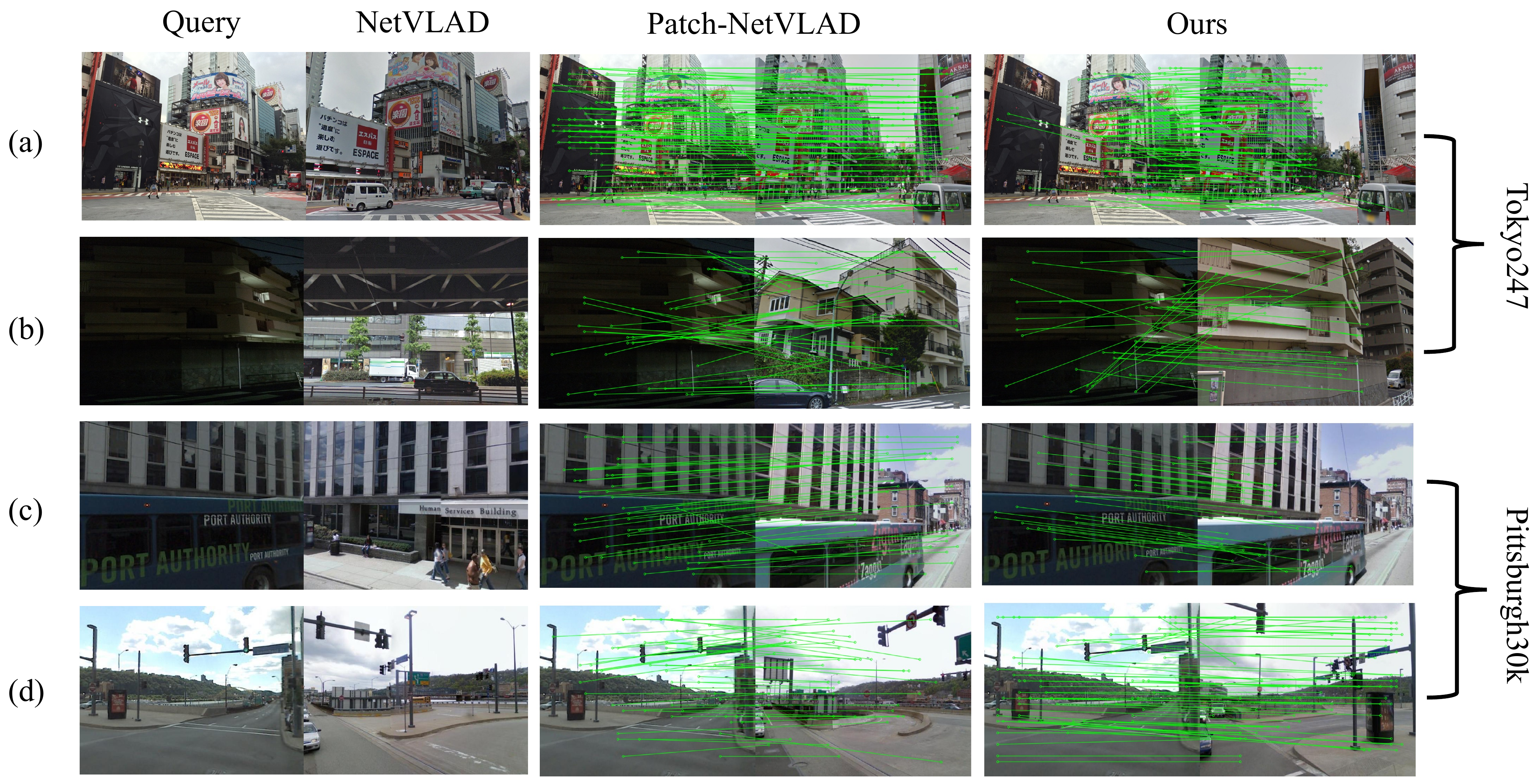}
  \caption{\textbf{Qualitative results.} 
  (a) and (c) are successfully retrieved by all methods, where green lines represent the matched patches. 
  (b) and (d) are successfully retrieved by Patch-NetVLAD+ while NetVLAD and Patch-NetVLAD produce an incorrect result.
  It can be seen that Patch-NetVLAD+ is capable to match patches in LSR of the image, such as buildings and billboards.
  }
  \label{fig_quantitative}
\end{figure*}

We compare with classic NetVLAD\cite{arandjelovic2016NetVLAD} and  state-of-the-art patch-based VPR methods Patch-NetVLAD\cite{hausler2021patch} in the test-set of Pittsburgh30k and Tokyo247 in this experiment.
Our method is only fine-tuned on the Pittsburgh30k train-set.
In pairwise image matching, we use the same $k=100$ candidate images for Patch-NetVLAD and Patch-NetVLAD+.

Quantitative comparisons of NetVLAD, Patch-NetVLAD and Patch-NetVLAD+ are shown in \Reftab{tb_quantitative}.
Patch-NetVLAD+ achieves 88.50\% in Recall@1 on Pittsburgh30k, better than the 87.40\% achieved by Patch-NetVLAD with an improvement of 1.1\%, and our method outperforms NetVLAD by 4.84\%.
The difference is particularly noticeable in the challenging Tokyo247 dataset, where the Recall@1 of our method is significantly improved to 83.49\% compared to 77.14\% achieved by Patch-NetVLAD.
And our method achieves up to 16.51\% performance improvement against NetVLAD.
In addition to Recall@1, our method outperforms the baseline on both Recall@5 and Recall@10.

Furthermore, the descriptor dimension can be reduced using PCA. 
We experimented in both 512 and 4096 dimensions (\ie, $D_{pca}=512,4096$).
The results obtained are consistent.
As shown in \Reffig{fig_compare}, for both the Tokyo247 dataset and the Pittsburgh30k dataset, our method outperformed the baseline methods for different recall rates, and in particular, on the Tokyo247 dataset, our method is substantially ahead of baseline methods.

The quantitative comparison of our method with baseline is shown in \Reffig{fig_quantitative}.
In \Reffig{fig_quantitative}(a)(c) all three methods were successed.
However, Patch-NetVLAD shows many mismatched patches that are located in non-LSR of the image, such as the sky.
Our method produced much fewer mismatched patches due to the weighted matching strategy, and most of the matched patches were located in the LSR of the image, such as buildings and billboards.
As seen in \Reffig{fig_quantitative}(b)(d), in difficult scenes, our method can still successfully retrieve the correct results by matching patches the in LSR of the image.
Such performance validates the effectiveness of the fine-tuning strategy and the weighted matching strategy.

\subsection{Ablation Studies}

\setlength{\tabcolsep}{14pt} 
\begin{table*}[tbp]
  \small
  \begin{center}
      \caption{Ablation studies} 
      \label{tb_ablation}
      \begin{tabular}{@{}c|ccc||ccc@{}}
      \toprule
      \multirow{2}{*}{Method ($D_{pca}=4096$)}          & \multicolumn{3}{c||}{Pittsburgh30k} & \multicolumn{3}{c}{Tokyo247} \\ \cmidrule(l){2-7} 
                                                       & R@1         & R@5        & R@10        & R@1     & R@5    & R@10  \\ \midrule \midrule 
      Baseline (Patch-NetVLAD\cite{hausler2021patch})  & 87.40       & 93.66      & 95.36       & 77.14   & 84.44  & 86.98 \\
      Ours w/o fine-tune                               & 87.19       & 93.78      & 95.63       & 81.90   & 86.35  & 88.25 \\
      Ours w/o weighted match                          & 88.45       & 94.32      & 95.73       & 79.05   & 87.30  & 89.52 \\ \hline
      Ours                                             & 88.50       & 94.47      & 95.79       & 83.49   & 87.94  & 90.16 \\ \bottomrule
      \end{tabular}
  \end{center}
  \vspace{-0.3cm}
\end{table*}

\begin{figure*}[t]
  \centering
      $\begin{array} {cc}
      \includegraphics[width=3in]{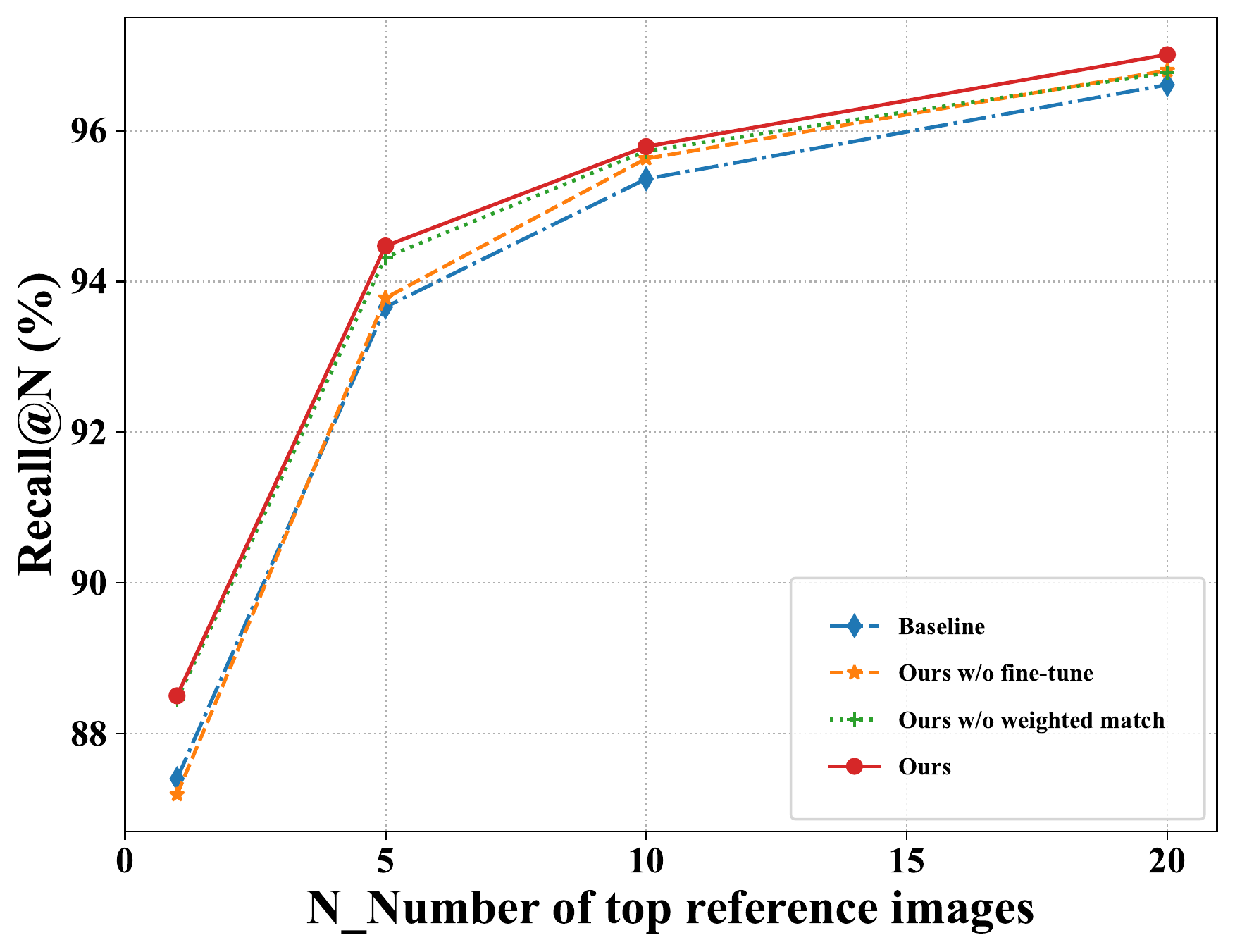} &
      \includegraphics[width=3in]{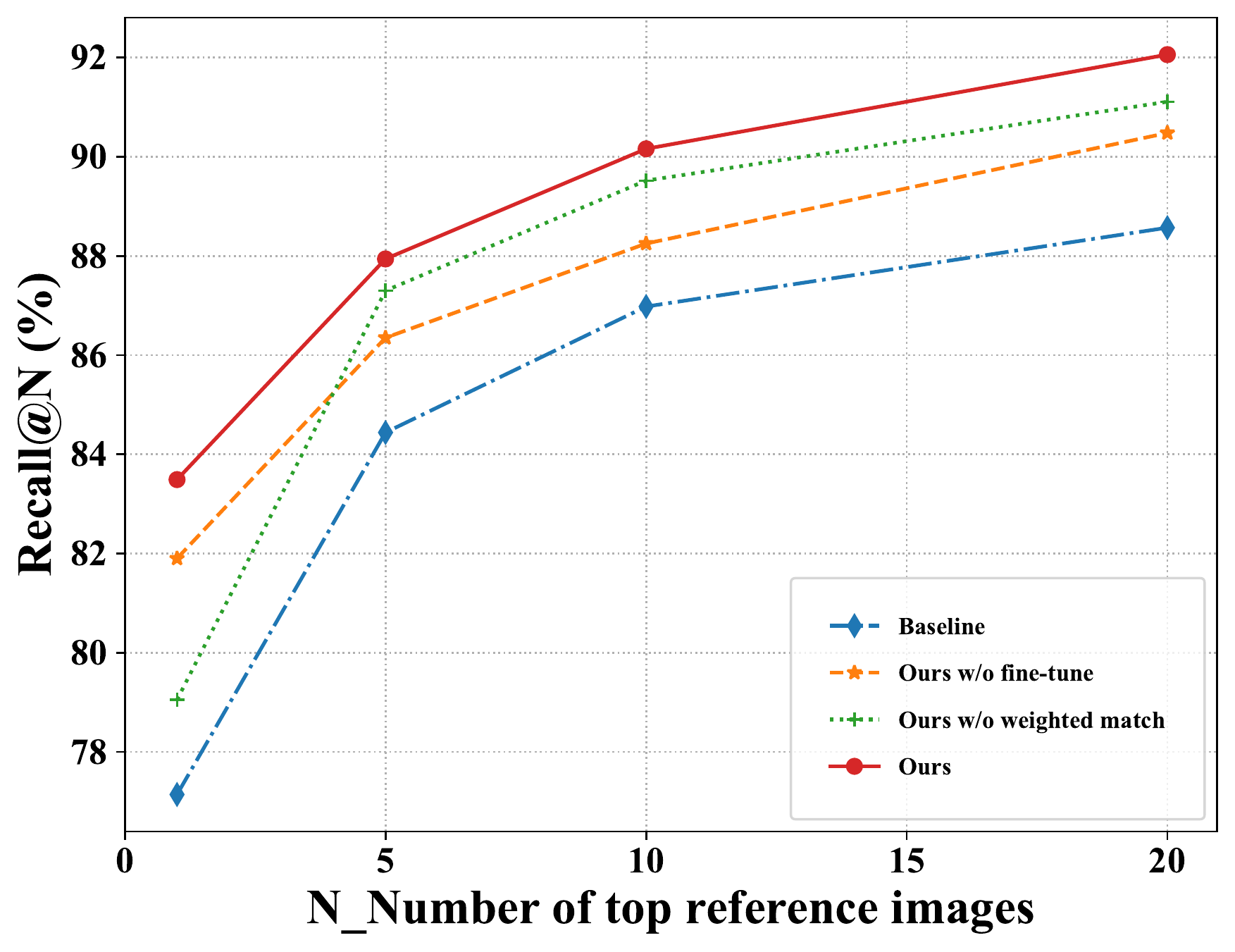} \\
      \mbox{(a) Pittsburgh30k} &\mbox{(b) Tokyo247}
      \end{array}$
  \caption{\textbf{Ablation Studies.} We show the Recall@N performs of ablation studies on Pittsburgh30k and Tokyo247. (1) \emph{Baseline}: Patch-NetVLAD; (2) \emph{Ours w/o fine-tune}: Patch-NetVLAD+ without fine-tuning strategy; (3) \emph{Ours w/o weighted}: Patch-NetVLAD+ without weighed matching strategy; (4) \emph{Ours}:Patch-NetVLAD+ with fine-tuning strategy and weighed matching strategy.
  }
  \label{fig_ablation}
\end{figure*}

We conducted ablation studies on Pittsburgh30k and Tokyo247 to analyze the effectiveness of the \emph{fine-tuning strategy} and the \emph{weighed matching strategy}.
In the ablation studies, the baseline is Patch-NetVLAD with \emph{patch size $=5$} and \emph{Rapid Spatial Scoring}, all methods use 4096-dimensional descriptors.
We set up a total of four methods: (1) \emph{Baseline}: Patch-NetVLAD; (2) \emph{Ours w/o fine-tune}: Patch-NetVLAD+ without fine-tuning strategy; (3) \emph{Ours w/o weighted}: Patch-NetVLAD+ without weighed matching strategy; (4) \emph{Ours}: Patch-NetVLAD+ with fine-tuning strategy and weighed matching strategy.

\subsubsection{Effectiveness of fine-tuning the NetVLAD}
\label{sec_effect_finetune}
The baseline method directly utilized the original NetVLAD, which was trained using the whole images of Pittsburgh30k to extract descriptors for patches. 
Our method uses the Pittsburgh30k dataset to fine-tune the original NetVLAD to enable it to extract patch-level descriptors.
As shown in \Reftab{tb_ablation}, in the Pittsburgh30k dataset, our method achieves 1.05\% (Recall@1) improvement with the fine-tuning strategy compared to the baseline, and a 1.31\% (Recall@1) decrease after removing the fine-tuning.
The effect of the fine-tuning strategy was more pronounced in the Tokyo247 dataset, where the changes expanded to 1.91\% and 1.59\% respectively.
The results of the two comparison experiments clearly show the effectiveness of the fine-tuning strategy.

\subsubsection{Effectiveness of weighted matching strategy}
Our approach assigns weights to patches and updates the distance matrix of pairwise matching by the weights of patches to implement the weighted matching strategy.
As a result, patches in LSR play a greater role in matching.
In the Pittsburgh30k dataset, the role of the weighted matching strategy is not obvious in Recall@1, leading the baseline approach in Recall@5 and Recall@10, as shown in \Reftab{tb_ablation} and \Reffig{fig_ablation} (a).
In the Tokyo247 dataset, the weighted matching strategy was effective.
In (\emph{Baseline} vs \emph{Ours w/o fine-tune}) comparison,  Recall@1 is improved by 4.76\% and in (\emph{Ours w/o weighted match} vs \emph{Ours}) comparison, Recall@1 is improved by 4.44\%.

As shown in \Reffig{fig_ablation}, the result curves of the \emph{Baseline} are both lower than the result curves of \emph{Ours without fine-tuning} (green) and \emph{Ours without weighting} (yellow), illustrating the effectiveness of the fine-tuning and the weighted matching strategies.

\section{Conclusion}
In this paper, we propose a novel patch-based VPR method named Patch-NetVLAD+ which consists of a fine-tuning strategy and a weighted matching strategy.
The fine-tuning strategy is used to make original NetVLAD more suitable for extracting patch-level descriptors.
The weighted matching strategy is used to find patches in LSR and make these patches easy to match by assigning a large weight to them.
Quantitative and Qualitative experiments conducted on Pittsburgh30k and Tokyo247 have demonstrated the excellent performance of our method.
In the future, we intend to aggregate patch descriptors with the semantic contextual information as a global descriptor that can retrieve results with stable multiple local regions.

%


{\small
\bibliographystyle{ieee_fullname}
\bibliography{egbib}
}

\end{document}